\documentclass[runningheads]{llncs}

\usepackage{eccv}

\usepackage{eccvabbrv}

\usepackage{graphicx}
\usepackage{booktabs}

\usepackage[accsupp]{axessibility}  
\usepackage{marvosym}
\newcommand{\corrauth}{\textsuperscript{\Letter}}

\usepackage{hyperref}

\usepackage{orcidlink}
\usepackage{multirow}

\begin{document}

\title{Embedding Rotation Invariance for Provable Multi-Oriented Scene Text Recognition} 

\titlerunning{Embedding Rotation Invariance for Provable Multi-Oriented STR}

\author{Zhibin Ma\inst{1, 2} \and
Pengwen Dai\inst{1, 2}\corrauth \and
Yi Liu\inst{3} \and Xugong Qin\inst{4} \and Chenyun Yu\inst{1} \and Xiaochun Cao\inst{1, 2}}

\begingroup
\renewcommand{\thefootnote}{\Letter}
\footnotetext[1]{Corresponding author.}
\endgroup
\authorrunning{Z. Ma et al.}

\institute{Shenzhen Campus of Sun Yat-sen University, China \\
\and
Shenzhen Key Laboratory of Adversarial Artificial Intelligence, China\\
\and
Baidu Inc., China\\
\and
Nanjing University of Science and Technology, China\\
\email{mazhb3@mail2.sysu.edu.cn, \{daipw, yuchy35, caoxiaochun\}@mail.sysu.edu.cn, liuyi22@baidu.com, qinxugong@njust.edu.cn}\\
}
\maketitle

\begin{abstract}
  Multi-oriented text is ubiquitous in real-world scenes and remains a major challenge for scene text recognition (STR). Existing rotation-aware methods explicitly estimate text orientation. However, due to the lack of theoretical guarantees, they are prone to error accumulation, increased computational cost, and strong reliance on data. In this work, we incorporate rotation invariance into the STR framework to address these limitations.
  Specifically, we adopt an encoder--decoder architecture, embedding rotation equivariance in the encoder and rotation invariance in the decoder to construct a fully rotation-invariant network. On the decoder side, we first identify and prove the rotation-invariant property of the cross-attention mechanism and use it to formulate a rotation-invariant text decoder that maps visual features to output text in a rotation-invariant manner. On the encoder side, we propose a rotation-equivariant local–global extraction network that integrates deep equivariant convolutions with self-attention, enabling rotation-equivariant feature extraction while modeling inter-character dependencies and preserving fine-grained visual details.
  By integrating the encoder and decoder, we obtain an end-to-end \textbf{R}otation-\textbf{I}nvariant \textbf{S}cene \textbf{TE}xt \textbf{R}ecognition network (RISTER). 
  RISTER provides rotation invariance with theoretical guarantees, enhancing robustness on multi-oriented samples without introducing additional inference computation or relying on data-driven orientation correction. Experiments show that RISTER achieves state-of-the-art performance on both standard and multi-oriented benchmarks, surpassing the second-best model by 4.0\% in accuracy on the general multi-oriented dataset.
\end{abstract}

\section{Introduction}
\label{sec:intro}

\begin{figure}[t!]
    \centering
    \includegraphics[width=1.0\textwidth]{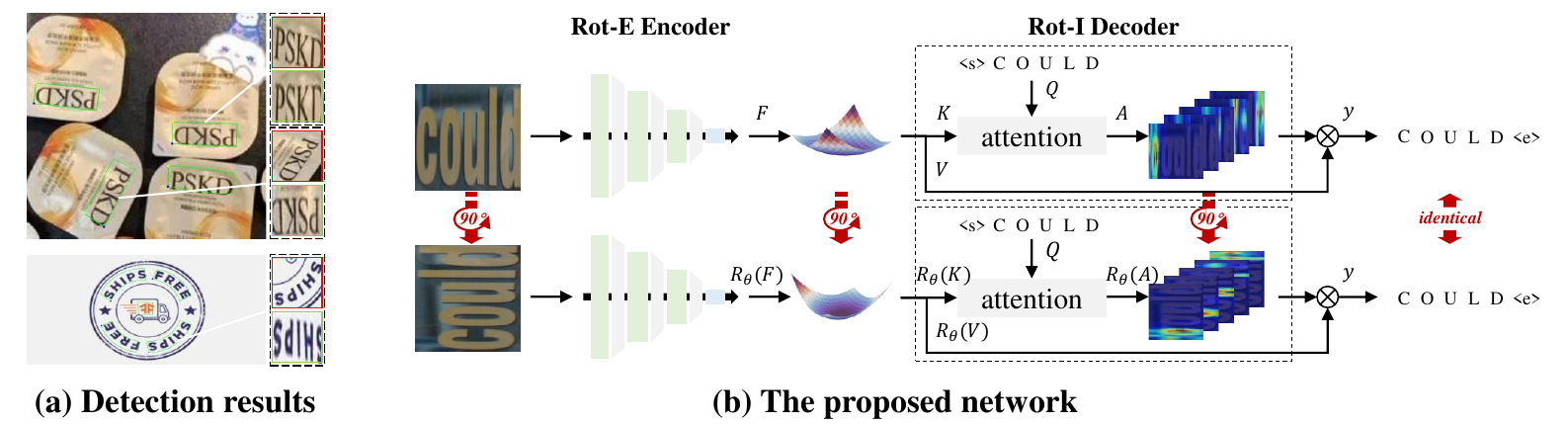}
    \caption{(a) The detection results produced by LRANet~\cite{aaai24/lranet/YCSu}. The red boxes indicate results obtained by cropping with the minimum enclosing bounding boxes (arbitrary angles), while the green boxes indicate results obtained using TPS-based rectification (canonical angles).  (b) In our proposed rotation-invariant network, rotating the input image induces equivariant transformations in both visual features and attention maps. After the attend operation, they lead to exactly the same prediction.}
    \label{fig1}
\end{figure}
Scene Text Recognition (STR) aims to translate the text in images into a machine-readable form. Its wide applications in areas such as autonomous driving~\cite{tai22/autodriving/CSZhang}, visual understanding~\cite{lumos}, and embodied intelligence~\cite{icirs10/robot/Posner} have made it a popular research topic.
Existing STR methods primarily focus on enhancing the model’s language modeling capability, either by introducing explicit language models~\cite{cvpr21/abinet/SCFang,aaai23/trocr/MHLi, eccv22/parseq/Bautista, cvpr20/srn/DLYu} or by integrating visual feature extraction and language modeling into a unified framework~\cite{arxiv/svtrv2/YKDu, aaai24/busnet/JJWei,ijcai22/svtr/YKDu}. 
However, the orientation of text in images, which is one of the most common and important issues, has been largely overlooked~\cite{iccv23/maerec/QJiang}. In real-world scenarios, text appears in various orientations. While most popular text detectors regress boundary points to obtain tightly fitted text bounding boxes, they inevitably produce multi-oriented text instances, primarily with orientations of 0°, 90°, 180°, and 270°, as shown in Figure~\ref{fig1} (a). These samples, referred to as canonical-angle samples in this paper, pose significant challenges to the recognition task, while existing STR methods typically assume that the text in input images is arranged from left to right. When the text orientation deviates significantly from the horizontal direction, existing text recognizers are prone to producing incorrect outputs~\cite{iccv23/maerec/QJiang}. 

Although a limited number of studies have focused on recognizing multi-oriented text, they all face the following issues. (1) No theoretical guarantees: Methods based on multi-oriented encoding~\cite{AON} or rectification modules~\cite{tpami19/aster/BGShi, esir, arxiv/svtrv2/YKDu} require the model to perceive text orientation explicitly. If the orientation perception network (e.g., the localization network in ASTER~\cite{tpami19/aster/BGShi}) fails to estimate the text orientation correctly, the model struggles to recognize the text.
(2) More computational cost: The introduction of an additional perception network incurs extra computation, leading to reduced inference efficiency.
(3) Reliance on data augmentation: Rotation-based data augmentation during training becomes necessary, forcing the model to separately learn features for different orientations, which reduces feature utilization efficiency.

Inspired by existing rotation-equivariant networks~\cite{groupConv, FConv, generalization}, we attempt to provide a unified solution to the above problems by embedding rotation equivariance into STR networks. However, current rotation-equivariant networks are not directly suitable for STR, as they are primarily designed for classification~\cite{groupConv} or low-level vision tasks~\cite{imgfusion,imgrecons}. In contrast, STR is a unique variable-length, multi-label recognition problem, which, unlike other tasks, requires achieving rotation invariance in an end-to-end manner rather than mere equivariance.

In this work, we incorporate rotation invariance into the STR network to address the aforementioned issues. Specifically, we identify a previously overlooked rotation-invariant property of the cross-attention mechanism shared by a class of attention-based STR decoders. Our study demonstrates that the cross-attention mechanism exhibits a rotation-invariant property when operating on 2d feature inputs, namely, the attended features remain identical before and after arbitrary spatial rotations of the visual features. We further analyze this distinctive property and provide a theoretical proof. Based on this observation, we derive a set of design principles for constructing \textit{Rotation-Invariant Text Decoder (RITD)}. By minimally modifying existing decoder architectures, RITD realizes a rotation-invariant mapping from visual features to the text output space, producing consistent variable-length predictions and confidence scores under input rotations (exact at canonical orientations and approximate at other orientations).


While RITD ensures rotation-invariant decoding, achieving an end-to-end rotation-invariant STR system further requires rotation-equivariant visual feature extraction. This motivates us to combine RITD with rotation-equivariant visual encoders. However, existing rotation-equivariant backbones are not well-suited for STR. Specifically, purely convolutional designs~\cite{groupConv, FConv, pdoeconv} fail to capture inter-character dependencies, while attention-based counterparts~\cite{stand-alone, e2equvit, lietransformer} typically rely on shallow convolutional tokenization, leading to early downsampling and the loss of fine-grained details. To address these issues, we propose a hierarchical \textit{Rotation-Equivariant Local–Global Extraction (RELG)} network that integrates deep equivariant convolutions with self-attention, enabling rotation-equivariant feature extraction while preserving detailed geometric information and modeling inter-character dependencies and encodes positional information. 
Finally, by combining RELG and RITD into an encoder-decoder network, we construct a \textit{\textbf{R}otation-\textbf{I}nvariant S\textbf{TR} network (RISTER)}. The inherent rotation invariance of RISTER ensures identical recognition results before and after rotations of the input image (As shown in Figure~\ref{fig1} (b)). This property allows RISTER to recognize multi-oriented text without relying on any correction modules or data augmentation strategies, maintaining stability across various orientations while avoiding additional computational overhead. Furthermore, integrating rotation invariance boosts data utilization and strengthens the aggregation of key features, thereby improving the model’s robustness across diverse and challenging benchmarks.
Our contributions are summarized as follows:
\begin{itemize}
    \item We incorporate rotation invariance into STR, which leads to significantly improved robustness to multi-oriented text and mitigates the computational and data-efficiency limitations of prior rotation-aware methods.
    \item A theoretical proof of the rotation-invariant property of cross-attention is provided. Based on this analysis, we derive the design principles for constructing \textit{Rotation-Invariant Text Decoder} (RITD) that establishes a rotation-invariant mapping from the feature space to the output text space.
    \item A \textit{Rotation-Equivariant Local–Global Extraction} (RELG) network is proposed as the encoder backbone, characterized by learning a rotation-equivariant image-to-feature mapping while capturing inter-character dependencies and preserving fine-grained visual details.
    \item By combining RELG with RITD, a \textit{Rotation-Invariant STR network} is built, which exhibits identical recognition results before and after image rotation. Experimental results demonstrate that RISTER consistently achieves state-of-the-art performance on both standard and multi-oriented benchmarks.
\end{itemize}

\section{Related Works}
\subsection{Multi-oriented Text Recognition}
Existing STR methods typically recognize multi-oriented text by combining 2d feature extraction with attention mechanisms~\cite{aaai19/sar/HLi, eccv22/matrn/Na, iccv23/maerec/QJiang, eccv22/parseq/Bautista, tpami25/igtr/YKDu}. Unlike 1d feature extractors~\cite{cvpr16/rare/BGShi, tpami19/aster/BGShi, eccv20/robustscanner/XYYue} or CTC-based methods~\cite{tpami17/crnn/BGShi, ijcai22/svtr/YKDu, arxiv/svtrv2/YKDu}, these models retain spatial structures and use attention during decoding to localize and identify each character, enhancing recognition of irregular text layouts. However, their performance is highly dependent on data augmentation and the precision of the attention mechanism. Once attention fails, accuracy drops sharply.
Besides, AON~\cite{AON} encodes images from four directions to capture directional features and fuses them via a filter gating module, which increases computational cost and may introduce inconsistent spatial semantics across directions.
SLOAN~\cite{SLOAN} transforms images from Cartesian to polar coordinates, converting rotations into translations to leverage CNNs’ translation equivariance. Yet, the polar transformation causes spatial distortion that weakens feature extraction.
Other approaches, such as ASTER~\cite{tpami19/aster/BGShi}, ESIR~\cite{esir}, and SVTRv2~\cite{arxiv/svtrv2/YKDu}, attempt to rectify rotated text to a horizontal orientation via interpolation or feature rearrangement. These methods add extra computation and lack theoretical guarantees, leading to unstable performance on multi-oriented samples.

\subsection{Rotation Equivariant Deep Networks}
Incorporating rotation equivariance into neural networks has become a prominent research topic in computer vision. G-CNN~\cite{groupConv} first introduced a theoretical framework for equivariant convolutions under \text{$\pi$/2} rotations, inspiring a variety of Rot-E approaches~\cite{pdoeconv, FConv, hexaconv, pdoes2nn}. Methods such as~\cite{e2ecnn, steerconv} achieved continuous rotation equivariance through filter parameterization, while~\cite{pdoeconv, pdoes2nn} linked convolutions with differential operators for formal error analysis. Subsequent works extended equivariance to transformers~\cite{effieqinet, stand-alone, e2equvit, lietransformer}, and theoretical studies demonstrated that group-equivariant networks exhibit improved generalization capabilities~\cite{generalization}.
In applications, F-Conv~\cite{FConv} introduced Fourier-based filters to maintain rotational consistency in the super-resolution task, with subsequent works confirming enhanced stability in related vision tasks such as image inpainting~\cite{imginpanting}, image fusion~\cite{imgfusion}, or reconstruction~\cite{imgrecons}.

Despite these advances, rotation equivariant networks remain unexplored in STR. Unlike classification or low-level tasks, STR is a variable-length, mult-label sequence generation task, which requires rotation equivariance during feature extraction but rotation invariance during decoding to ensure orientation-independent semantic output. In addition, the unique requirements of STR tasks for backbones to model inter-character dependencies and perform progressive downsampling makes existing equivariant networks unsuitable for direct use as encoders. Therefore, achieving efficient equivariant feature extraction along with rotation-invariant decoding is the main focus of this work.

\section{Rotation-Invariant Text Recognition Network}
In this section, we first present the overall architecture of RISTER and describe how it realizes a rotation-invariant mapping from the input image to the output text. We then detail the architecture of the proposed Rotation-Equivariant Local–Global Extraction network. Finally, we introduce the rotation-invariant property of cross-attention and demonstrate how it can be exploited to achieve rotation-invariant variable-length decoding.

\subsection{Overall Architecture}
RISTER follows the widely adopted encoder--decoder paradigm in STR. In particular, we achieve an overall rotation-invariant architecture by embedding rotation equivariance into the encoder and rotation invariance into the decoder, respectively.
During visual encoding, the input image $x$ is fed into the
\textit{Rotation-Equivariant Local--Global Extractor (RELG)}
to perform feature extraction and sequence relationship modeling,
producing the visual representation $F$:
\begin{equation}
    F = RELG (x)\in \mathbb{R}^{c \times \frac{h}{8} \times \frac{w}{8}},
\end{equation}

where $c$ denotes the feature dimension, while $h$ and $w$ represent the height and width of the input image, respectively.
During decoding, $F$ together with the previously decoded sequence $V$
is input into the \textit{Rotation-Invariant Text Decoder (RITD)},
which autoregressively generates the complete text sequence $y\in \mathbb{R}^{L}$:
\begin{equation}
    y^t = RITD (F, V^{0:t-1})\in \mathbb{R}^{1},
\end{equation}
\begin{equation}
    V^t = Embedding (y^t)\in \mathbb{R}^{c \times 1},
\end{equation}

where $t = 1, \ldots, L$ denotes the time step in the autoregressive decoding process, $y^{t}$ represents the character decoded at time step $t$, and $y^{0}$ denotes the start-of-sequence token.

\subsection{Rotation-Equivariant Local-Global Extractor}
\begin{figure*}[t]
    \centering
    \includegraphics[width=1.0\textwidth]{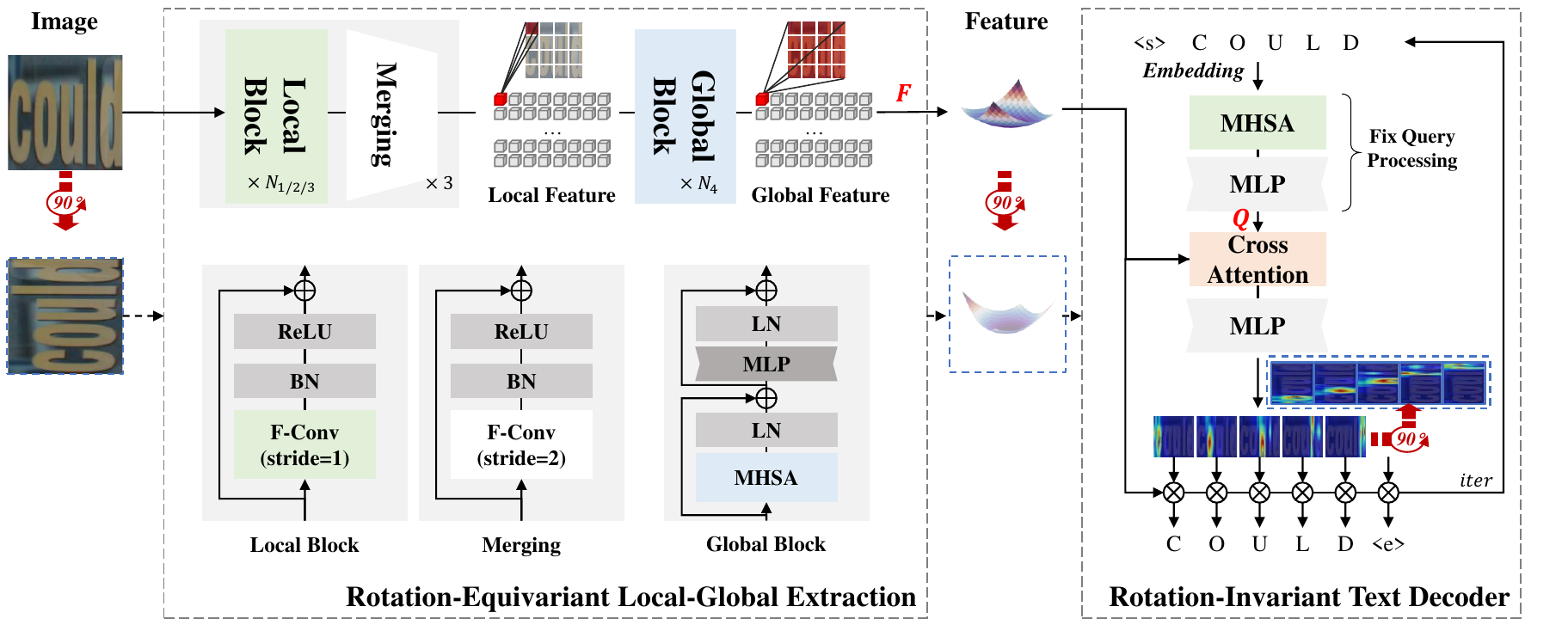}
    \caption{An overview of RISTER. It employs RELG for feature representation learning and decodes character sequences from the resulting visual features using RITD. When the image undergoes spatial rotation, we obtain equivariant visual features and attention maps (as shown in the blue dash boxes); after being attended in the decoder, the model produces rotation-invariant predictions.} 
    \label{fig2}
\end{figure*}
The RELG is designed with three objectives:
(1) to achieve rotation-equivariant visual feature extraction;
(2) to preserve fine-grained local features through progressive downsampling; and
(3) to model inter-character dependencies and positional information.
Although existing Rot-E networks~\cite{FConv, steerconv, pdoes2nn, stand-alone, lietransformer, e2equvit} can satisfy the first objective, they cannot simultaneously fulfill the latter two objectives, which limits their applicability to text recognition tasks.
In RELG, we address these two objectives through Rot-E local extraction and Rot-E global extraction, respectively.

\subsubsection{Rot-E Local Extraction}
We construct the Local Block using F-Conv~\cite{FConv}, a type of group-equivariant convolution, to enable rotation-equivariant local feature extraction while maximizing the preservation of fine-grained features through progressive downsampling. Specifically, RELG consists of three stages. In each stage, several local blocks are first applied for feature extraction, followed by a merging operation for spatial downsampling and channel expansion:
\begin{equation}
    F_k = Merging(\mathcal{LB}(F_{k-1}))\in \mathbb{R}^{c_k \times \frac{w}{2^k} \times \frac{h}{2^k}}, k = 1,2,3
\end{equation}
where $F_k$ denotes the feature output of the $k$-th stage, and $F_0=x$. $\mathcal{LB}(\cdot)$ is composed of $N_k$ stacked local blocks. In each local block, we employ F-Conv to perform local feature extraction. The convolution kernels are of size $3 \times 3$ with a stride of 1. 
For the choice of the rotation group, we adopt the discrete rotation group $C_4$ as~\cite{FConv} does, corresponding to rotations by canonical angles.  The merging operation is similar to a local block, except that it uses a stride of 2 and increases the number of output channels.
After each stage, the resolution of the input image is reduced to $1/2$, 
$1/4$, and $1/8$ of the original, respectively, while the number 
of channels is expanded to 96, 192, and 384.

\subsubsection{Rot-E Global Extraction}
Attention-based STR networks require visual features to encode inter-character dependencies, which necessitates global relationship modeling over these features.
We achieve this through a self-attention mechanism.
Notably, existing works have already demonstrated that self-attention layers exhibit equivariance to rotations~\cite{e2equvit, stand-alone}.
Specifically, global blocks take the visual feature $F_3$ as input and finally outputs the enhanced visual feature representation $F$:
\begin{equation}
F = \mathcal{GB}(F_3)\in \mathbb{R}^{c_3 \times \frac{w}{8} \times \frac{h}{8}},
\end{equation}
where $\mathcal{GB}(\cdot)$ denotes the stack of global blocks composed of multi-head self-attention and MLP layers.

\subsection{Design principles of Rotation-Invariant Text Decoder}
\begin{figure*}[t]
    \centering
    \includegraphics[width=1.0\textwidth]{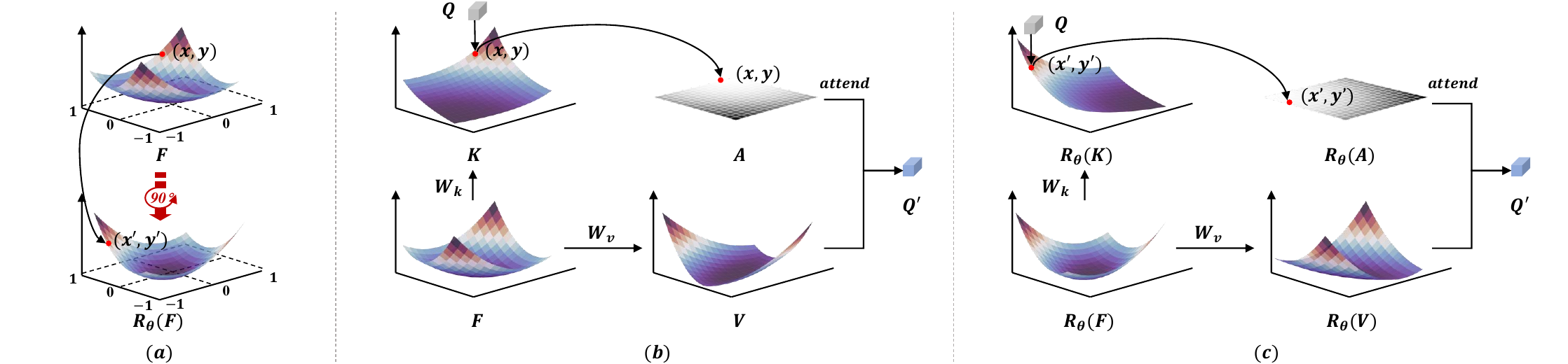}
    \caption{Illustration of the rotation invariant of cross-attention. Keeping the query token $Q$ fixed, the attended feature $Q'$ is identical before and after spatial rotation of $F$.}
    \label{fig3}
\end{figure*}
Rotation-invariant variable-length decoding is a core component of RISTER, achieved by exploiting the rotation invariance of cross-attention. We first describe the rotation-invariant property of cross-attention. Next, we outline the principles for constructing a rotation-invariant decoder, and finally present an autoregressive decoding architecture that realizes these principles.

\subsubsection{Rotation invariant of cross attention.}
Figure~\ref{fig3} illustrates the rotation-invariant property of cross-attention, which can be described as:
\begin{equation}
\mathcal{C}(Q, R_{\theta}(F)) \equiv \mathcal{C}(Q, F),
\label{cross-attention equal}
\end{equation}
where $\mathcal{C}$ denotes the cross-attention operation, $Q$ denotes the query tokens, $F$ denotes the source tokens, and $R_{\theta}$ denotes a spatial rotation of $F$ by an angle $\theta$. As illustrated in Figure~\ref{fig3}, with fixed query tokens, a spatial rotation applied to
$F$ results in equivariant transformations of the corresponding keys $K$,
values $V$, and attention map $A$, thereby guaranteeing that the
attended output remains rotation-invariant. The mathematical proof of Equation~\ref{cross-attention equal} is provided in Appendix A, where it is shown to hold exactly at canonical angles and approximately at other orientations.

\subsubsection{Design of the proposed decoder.} The validity of Eq.~\ref{cross-attention equal} reveals a fundamental property of cross-attention: it defines a mapping over unordered feature sets and is inherently insensitive to geometric structures. Therefore, as long as this invariance is carefully preserved, a decoder constructed using cross-attention can naturally enable rotation-invariant variable-length decoding. The design of RITD must satisfy the following two principles: (1) cross-attention is used to perform modality interaction, and (2) the query tokens fed into the cross-attention remain invariant under image rotation. Following these two principles, we adopt an autoregressive decoding scheme to implement our RITD, as illustrated in the right part of Figure~\ref{fig2}. Specifically, at time step $t$, we first establish dependencies among previously decoded characters via fixed query processing, thereby obtaining the query $Q$:
\begin{equation}
Q_{0:t-1} = MLP(MHSA(Embedding(y_{0:t-1}))),
\end{equation}
where $y_{0:t-1}$ denotes the previously decoded characters, and $y_0$ represents the start-of-sequence token.
Subsequently, $Q_{t-1}$ is used as the query token, and the visual features $F$ are treated as source tokens and fed into the cross-attention module to compute the attended output. After passing through an MLP for feature mapping, the predicted character is obtained by applying the $argmax$ operator over the output logits:
\begin{equation}
y_t= argmax(MLP(\mathcal{C}(Q_{t-1}, F))).
\end{equation}
Subsequently, $y_t$ is used as a decoded character to compute the query token for the next time step, and the decoder iteratively outputs the subsequent predicted characters. Since the query token $Q$ fed into the cross-attention remains invariant under image rotation, the decoder consistently produces identical predictions at the same time step before and after feature rotation.

\subsection{Optimization Objective}
The training objective follows the traditional AR Loss, formulated as:
\begin{equation}
\min_{\mathcal{W}} \; \mathbb{E}_{(x, y) \sim \mathcal{D}} \left[ - \sum_{t=1}^{L} \log p_{\mathcal{W}}\left( y_{t} \mid y_{<t}, x \right) \right],
\end{equation}
where $\mathcal{W}$ denotes the model parameters and $L$ is the maximum decoding length.

\subsection{Architecture Variants}
We design different variants of RISTER by adjusting its hyperparameters, including the number of local blocks and global blocks at each stage, as well as the number of attention heads in both the encoder and decoder. Detailed configurations are shown in Table~\ref{tab:archi}. By designing different RISTER variants, flexible choices can be offered for different application scenarios.

\begin{table}[h]
\centering
\caption{Architecture specifications of RISTER variants. \textit{Heads} denote the number of self-attention heads in both the global blocks/decoder. Measured on the IC13 dataset.
}
\begin{tabular}{c|c|c|c|c|c}
\hline 
  & $\left[N_1, N_2, N_3, N_4 \right]$ & \textit{Heads}  & FLOPs $(\times 10^9)$ & \textit{fps} &Params $(\times 10^6)$\\
\hline
Tiny  & {[}3,3,3,6{]}  & 6 & 7.22   & 42.1 & 15.9\\
Small  & {[}9,9,9,9{]}  & 6 & 12.99  & 31.4 &21.8\\
Base  & {[}12,12,12,15{]} & 12& 18.48  & 26.8 &32.8\\
Large  & {[}18,18,18,18{]} & 12& 24.25  & 22.7 &38.6\\
\bottomrule
\end{tabular}
\label{tab:archi}
\end{table}

\section{Experiments}
\subsection{Datasets and Implementation Details}
For training, we adopt the large-scale real-world dataset Union14M-Filter~\cite{arxiv/svtrv2/YKDu}, a refined version of Union14M-L~\cite{iccv23/maerec/QJiang}, which contains 3.2M training images from 17 datasets.
For evaluation, we test our method on 14 English benchmarks, including:  
(1) six common STR benchmarks (\textbf{CoB}): IIIT5k (3000)~\cite{bmvc12/iiit5k/Mishra}, SVT (647)~\cite{iccv11/svt/KWang}, IC13 (857)~\cite{icdar13/ic13/Karatzas}, IC15 (1811)~\cite{icdar15/ic15/Karatzas}, SVTP (645)~\cite{iccv13/svtp/Phan}, and CUTE80 (288)~\cite{esa14/cute80/Risnumawan};  
(2) \textbf{Union14M-benchmarks (U14M-B)}~\cite{iccv23/maerec/QJiang}, covering seven challenging categories: Curved (2426), Multi-Oriented (1369), Artistic (900), Contextless (779), Salient (1585), Multi-Words (829), and General (400k);  
(3) \textbf{ASOT} (3000)~\cite{SLOAN}, consisting of 3,000 text images with diverse orientations and lengths.  

During training, we use AdamW~\cite{iclr19/adamw/Loshchilov} with weight decay of $0.05$. The learning rate is set to $3.2 \times 10^{-5}$ for RISTER-T / S and $1.6 \times 10^{-5}$ for RISTER-B / L, with a batch size of $512$. A OneCycleLR scheduler with $1.5$ epochs of linear warm-up is applied over $20$ epochs. All images are resized to $128 \times 128$. Owing to the intrinsic rotation invariance of our method, rotation-based augmentation is omitted, whereas the remaining data augmentation settings follow those of PARSeq~\cite{eccv22/parseq/Bautista}. The maximum text length $L$ is $25$, and the character set size $N$ is $96$, including letters, digits, punctuation, and special tokens (start, end, pad). The results are evaluated using the Word Accuracy Ignore Cases (WAIC) metric. All experiments are conducted on a single $80$\,GB A100 GPU.


\subsection{Effectiveness of the proposed framework}
\begin{figure}[h]
    \centering
    \includegraphics[width=1.0\textwidth]{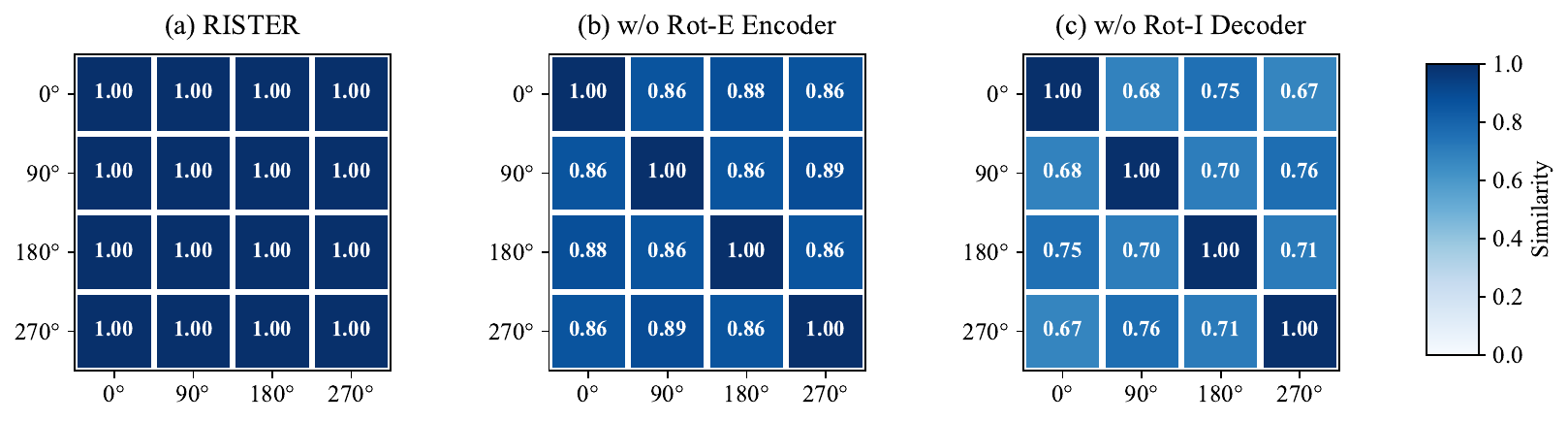}
    \caption{Cosine similarity between the output logits before and after image rotation. A higher similarity indicates greater consistency between the model outputs. A similarity of 1 means that the predictions and their corresponding confidence scores are exactly identical before and after image rotation. Measured on the IC13 dataset.}
    \label{fig4}
\end{figure}

RISTER exhibits end-to-end rotation invariance, from the input image to the final predicted text. To verify this property, we rotate the input images and measure the similarity between the logits produced by the model. As illustrated in Figure~\ref{fig4}(a), RISTER yields exactly the same logits for images before and after rotation, thereby confirming its strict rotation invariance under canonical-angle rotations. Consequently, the model demonstrates strong zero-shot rotation generalization, accurately recognizing images with diverse semantic orientations despite being trained exclusively on horizontally oriented images.
The rotation invariance of RISTER arises from the combination of a rotation-equivariant encoder and a rotation-invariant decoder. To verify the roles of these two components, we design two ablation variants:
(a) replacing the equivariant convolutions in RELG with standard convolutions, thereby breaking the rotation-equivariant property of the encoding process;
(b) replacing RITD with a CTC-based decoder~\cite{icml06/ctc/Graves, tpami17/crnn/BGShi}, thereby destroying the rotation-invariant property of the decoding process. As shown in Figure~\ref{fig4}(b, c), the absence of either the Rot-E encoder or the Rot-I decoder causes the model’s outputs to vary with the rotation of the input image, indicating that the model loses end-to-end rotation invariance. The quantitative results presented in Table~\ref{ablaRISTER} further corroborate the effectiveness of RISTER. Owing to its intrinsic rotation invariance, RISTER attains consistent recognition accuracy across images of varying orientations. This equivariant recognition capability enables robust performance on multi-oriented text. In contrast, the baseline models (a, b) exhibit a substantial decline in recognition accuracy when subjected to image rotations. We also experimented with incorporating random rotation augmentation during training, as adopted in AON~\cite{AON}. As shown in Table~\ref{ablaRISTER}, random rotation augmentation effectively improves the model’s performance on multi-oriented samples. However, with the model capacity fixed, reducing the proportion of horizontally oriented samples in the training data encourages the model to learn more orientation-invariant features, which in turn leads to a degradation in recognition accuracy on horizontal samples. In contrast, our RISTER achieves the highest recognition accuracy at every orientation. These observations suggest that incorporating rotation invariance into the model architecture is a more principled approach to multi-oriented recognition than relying on data-driven strategies.

\begin{table}[t]\footnotesize
\centering
\caption{Ablation study on the encoder–decoder framework. Measured on IC13.}

\setlength{\tabcolsep}{1.5pt}{
\begin{tabular}{l|ccccc}
\hline
{Method} 

&$0^\circ$ & $90^\circ$ & $180^\circ$ & $270^\circ$ & Avg.   \\

\hline
(a): w/o Rot-E Encoder &
\textbf{98.4} & 97.9 & 95.6 & 96.0 & 96.98 \\
(b): w/o Rot-I Decoder &
98.0 & 91.2 & 95.6 & 90.5 & 93.82 \\
(c): (a) + random-rotation &
97.5 & 96.9 & 97.7 & 96.9 &  97.25\\
(d): (b) + random-rotation &
97.3 & 93.6 & 97.9 & 93.1 &  95.48\\ 
(e): RISTER-S (Ours) &
\textbf{98.4} &\textbf{ 98.4} & \textbf{98.4} & \textbf{98.4} & \textbf{98.40} \\
\hline
\end{tabular}}

\label{ablaRISTER}
\end{table}

\subsection{Effectiveness of the proposed RELG}
RELG is an efficient rotation-equivariant backbone tailored for STR. To verify its effectiveness, we compare it with existing Rot-E backbone networks~\cite{pdoeconv,FConv,groupConv,stand-alone}. Since most existing Rot-E networks are primarily designed for classification tasks, their model scales differ by orders of magnitude from those commonly used in STR. To ensure a fair comparison, we incorporate their rotation-equivariant parameterized layers into the standard STR backbone ResNet-45~\cite{resnet} or ViT~\cite{ViT}, thereby constructing Rot-E encoders with comparable model capacity. All variants adopt the RITD design proposed in Section 3.3 as the decoder. As shown in Table~\ref{tab:encoder} (a-d), by incorporating Rot-E convolutions into existing convolution-based networks, the models exhibit rotation equivariance at the encoding stage. When combined with RITD, this property further enables end-to-end rotation invariance, resulting in consistent recognition performance across images with different orientations, and consequently improves the recognition accuracy on multi-oriented samples.
However, due to the lack of global modeling of visual features, purely convolutional backbones are not the optimal choice for STR. As shown in Table~\ref{tab:encoder} (e, h), backbones equipped with self-attention, such as ViT and SVTR, achieve significantly higher performance on images with 0° orientation compared to those in Table~\ref{tab:encoder} (a–d). This observation motivates us to incorporate attention mechanisms into the Rot-E backbone. Specifically, We reconstruct the ViT by leveraging existing group-equivariant self-attention methods~\cite{stand-alone, lietransformer}. As illustrated in Table~\ref{tab:encoder} (f, g), after introducing rotation equivariance into ViT, the models achieve improved recognition performance on multi-oriented samples; however, their accuracy on 0° samples decreases. We attribute this behavior to the fact that an overly shallow feature extraction process compresses the representational degrees of freedom of the model, while the early downsampling leads to the loss of local details, making it difficult for subsequent self-attention layers to recover the missing local geometric information.

\begin{table}[t]\footnotesize
\centering
\caption{Results of encoders comparison. $\dag$ indicates that the global blocks in RELG are replaced with an equal number of local blocks.}

\setlength{\tabcolsep}{0.5pt}{
\begin{tabular}{l|cccc|cccc|c|c}
\hline
{\multirow{2}{*}{Encoder}} &
\multicolumn{4}{c|}{IC13} &
\multicolumn{4}{c|}{SVT} &
\multirow{2}{*}{
    \begin{tabular}[c]{@{}c@{}}U14M-\\Mul-Ori\end{tabular}
} &
\multirow{2}{*}{
    \begin{tabular}[c]{@{}c@{}}Params\\ $(\times 10^6)$\end{tabular}
} \\

 &
$0^{\circ}$ & $90^{\circ}$ & $180^{\circ}$ & $270^{\circ}$ &
$0^{\circ}$ & $90^{\circ}$ & $180^{\circ}$ & $270^{\circ}$ &
 &  \\

\hline
(a): Resnet45 &
97.1 & 94.5 & 89.5 & 93.2 & 95.5 & 89.8 & 82.8 & 88.7 & 90.8  & 14.4  \\

(b): (a) + G-CNN~\cite{groupConv} &
97.2 & 97.2 & 97.2 & 97.2 & 94.9 & 94.9 & 94.9 & 94.9 & 90.9 & 14.4 \\
 
(c): (a) + PDO-eConv~\cite{pdoeconv} &
97.0 & 97.0 & 97.0 & 97.0 & 94.8 & 94.8 & 94.8 & 94.8 & 91.2 & 14.4  \\

(d): (a) + F-Conv~\cite{FConv} &
97.3 & 97.3 & 97.3 & 97.3 & 95.7 & 95.7 & 95.7 & 95.7 & 94.9 & 14.4  \\

(e): Vision Transformer~\cite{ViT} &
98.5 & 98.1 & 95.2 & 96.1 & 97.4 & 96.6 & 97.3 & 94.0 & 93.9 & 26.9 \\

(f): (e) + Stand-Alone~\cite{stand-alone} &
98.3 & 98.3 & 98.3 & 98.3 & 97.1 & 97.1 & 97.1 & 97.1 & 94.5 & 27.2 \\

(g): (e) + LieTransformer~\cite{lietransformer} &
97.9 & 97.9 & 97.9 & 97.9 & 96.4 & 96.4 & 96.4 & 96.4 & 94.0 & 28.5 \\

(h): SVTR~\cite{ijcai22/svtr/YKDu} &
\textbf{98.8} & 98.2 & 97.4 & 97.0 & 98.1 & 96.9 & 93.2 & 95.8 & 94.6 & 28.2  \\

(i): RELG$\dag$ &
98.0 & 98.0 & 98.0 & 98.0 & 96.4 & 96.4 & 96.4 & 96.4 & 95.0 &  28.4 \\

(j): RELG (Ours) &
98.6 & \textbf{98.6} & \textbf{98.6} & \textbf{98.6} & \textbf{98.3} & \textbf{98.3} & \textbf{98.3} & \textbf{98.3} & \textbf{96.6} & 32.8 \\

\hline
\end{tabular}}

\label{tab:encoder}
\end{table}

Unlike the aforementioned backbones, our proposed RELG adopts a deeper equivariant convolutional network and employs progressive downsampling, which preserves geometric details of the input images to the greatest extent. Meanwhile, the incorporation of self-attention enables the model to effectively capture intra-character patterns in text images. As a result, our model achieves strong performance on both 0° and multi-oriented samples. Compared with existing backbones, RELG attains an accuracy of 96.6\% on U14M-Multi-Oriented, outperforming the second-best encoder SVTR by 2\%, while achieving comparable recognition accuracy to SVTR on 0° samples. Furthermore, we validate the role of self-attention by replacing the Global Blocks in RELG with Local Blocks. As shown in Table~\ref{tab:encoder} (i), removing self-attention leads to degraded feature extraction capability, resulting in performance drops of 0.6\%, 1.9\%, and 1.6\% on IC13, SVT, and U14M-Multi-Oriented, respectively. These results further demonstrate the effectiveness of our design.

\subsection{Comparison of Computational Cost and Efficiency}
We analyze the computational efficiency of RISTER. Using group-equivariant convolutions (a) incurs the same inference cost as standard convolutions (b), because after training, the parameters are fixed through cyclic shifting, making the inference process identical to that of standard convolutions. Consequently, the FLOPs, parameter count, and FPS of RISTER are comparable to those of existing AR-based STR methods (e.g., NRTR), as reported in the Table~\ref{computation}. This demonstrates that RISTER achieves rotation-invariant recognition without introducing additional computational overhead, highlighting the advantage of our model in terms of inference efficiency.

\begin{table}[h]
\centering
\footnotesize
\setlength{\tabcolsep}{3pt}
\caption{Comparison of computational cost and inference speed. Measured on IC13.}
\begin{tabular}{l|ccc}
\hline
 & FLOPs $(\times 10^9)$ & Params $(\times 10^6)$ & FPS \\ 
\hline
(a) RISTER-S (with F-Conv) & 12.99 & 21.8 & 31.4\\ 
(b) RISTER-S (with standard conv.) & 12.99& 21.8 & 31.4 \\ 
(c) NRTR & 13.06 & 44.3 & 28.9 \\ 
\hline
\end{tabular}

\label{computation}
\end{table}

\subsection{Influence of Input Resolution}
Most STR models adopt an input resolution of 32×128, which results in visual feature maps with a height of only 4 pixels after feature extraction. When handling multi-oriented images, this leads to insufficient feature resolution for vertically oriented text, thereby degrading recognition accuracy~~\cite{iccv23/maerec/QJiang}. In contrast, methods designed for multi-oriented text recognition typically employ square input resolutions, with equal height and width (e.g., 100×100 in AON~\cite{AON}).

We experimentally investigate the impact of input resolution on model performance. As shown in Table~\ref{resolution}, we investigate the performance of two recent STR models, SVTRv2~\cite{arxiv/svtrv2/YKDu} and IGTR~\cite{tpami25/igtr/YKDu}, under different input resolutions. The experimental results indicate that increasing the resolution to 128×128 improves recognition performance on multi-oriented samples, while the recognition accuracy on horizontally oriented images remains almost unchanged. 
Therefore, we follow this setting and set the input resolution of RISTER to 128×128. Under the same input resolution, our RISTER still outperforms the latest models on multi-oriented samples by 6.00\% and 2.95\%  in accuracy, while achieving very comparable performance on common benchmarks. This demonstrates the superiority of integrating rotation invariance into the model.

\begin{table}[t]
\footnotesize
\centering
\caption{Model performance under different input resolutions}
\label{resolution}

\setlength{\tabcolsep}{1.5pt}
\begin{tabular}{l|cccc|ccc}
\hline
\multirow{2}{*}{Method (Resolution)} 
& \multicolumn{4}{c|}{Common} 
& \multicolumn{3}{c}{Oriented} \\
& IC13 & SVT & CUTE & Avg. 
& Mul-Ori & ASOT & Avg. \\
\hline
(a) SVTRv2 (32$\times$128) \cite{arxiv/svtrv2/YKDu} 
& 98.7 & 98.0 & 99.0 & 98.57 & 89.5 & 86.5 & 88.00  \\
(b) SVTRv2 (128$\times$128) \cite{arxiv/svtrv2/YKDu} 
& 98.6 & 98.1 & 98.8 & 98.50 & 90.2 & 87.1 & 88.65 \\
(c) IGTR (32$\times$128) \cite{tpami25/igtr/YKDu} 
& 98.6 & 98.3 & 97.6 & 98.17 & 92.6 & 90.1 & 91.35 \\
(d) IGTR (128$\times$128) \cite{tpami25/igtr/YKDu} 
& 98.6 & 98.4 & 97.7 & 98.20 & 92.9 & 90.5 & 91.70 \\
(e) RISTER-S (128$\times$128) &
98.6 & 98.3 & 98.3 & 98.40 & 96.0 & 93.3 & 94.65 \\
\hline
\end{tabular}
\end{table}

\subsection{Comparison with Oriented Text Recognizers}
Some existing STR methods are specifically designed for multi-oriented text, such as multi-orientation feature extraction~\cite{AON}, polar transformation~\cite{SLOAN}, or rectification modules~\cite{tpami19/aster/BGShi,arxiv/svtrv2/YKDu}. We compare these methods with our proposed RISTER, as shown in Table~\ref{tab:ori_methods}. The results show that on common samples, RISTER achieves recognition performance very close to the existing state-of-the-art SVTRv2~\cite{arxiv/svtrv2/YKDu}. On oriented samples, our method significantly outperforms existing approaches, achieving a 4.00\% improvement over the second-best method. This demonstrates that the proposed RISTER not only handles oriented samples effectively but also delivers impressive performance on common samples. 

\begin{table}[t]\footnotesize
\centering
\caption{Comparison results with Oriented Text Recognizers. 
}

\setlength{\tabcolsep}{1.5pt}{
\begin{tabular}{l|cccc|ccc}
\hline
\multirow{2}{*}{Method} &
\multicolumn{4}{c|}{Common} &
\multicolumn{3}{c}{Oriented} \\

&IC13 & SVT & CUTE & Avg. & Mul-Ori & ASOT & Avg.  \\

\hline

AON \cite{AON} &
91.5 & 82.8 & 76.8 & 83.70 & - & - & - \\

SLOAN \cite{SLOAN} &
96.0 & 96.1 & 92.4 & 94.83 & 91.1 & 90.2 & 90.65 \\
ASTER \cite{tpami19/aster/BGShi} &
96.2 & 94.6 & 93.4 & 94.80 & 78.8 & 80.8 & 79.80 \\
SVTRv2 \cite{arxiv/svtrv2/YKDu} &
\textbf{98.7} & 98.0 & \textbf{99.0} & \textbf{98.57} & 89.5 & 86.5 & 88.00 \\

\hline

RISTER-S&
98.6 & \textbf{98.3} & 98.3 & 98.40 & \textbf{96.0} & \textbf{93.3} & \textbf{94.65} \\

\hline
\end{tabular}}

\label{tab:ori_methods}
\end{table}

\begin{table*}[t]
\scriptsize
\centering
\caption{
Comparison on Union14M-Benchmarks \& Common Benchmarks.
\textbf{Bold} and \underline{underlined} values denote the 1st and 2nd results in each column.
}
\setlength{\tabcolsep}{0.5pt}
\renewcommand{\arraystretch}{1.1}

\begin{tabular}{l|ccccccc|cccccc|c|c}
\hline
\multirow{2}{*}{Method}
& \multicolumn{7}{c|}{Union14M-Benchmarks}
& \multicolumn{6}{c|}{Common Benchmarks}
& \multirow{2}{*}{Avg.}
& \multirow{2}{*}{
    \begin{tabular}[c]{@{}c@{}}Params\\ $(\times 10^6)$\end{tabular}
} \\

& CUR & MO & ART & CTL & SAL & MTW & GEN
& IC13 & SVT & IIIT & IC15 & SVTP & CUTE
& & \\

\hline
CRNN~\cite{tpami17/crnn/BGShi} 
& 19.4 & 4.5 & 34.2 & 44.0 & 16.7 & 35.7 & 60.4
& 91.8 & 83.8 & 90.8 & 71.8 & 70.4 & 80.9
& 54.18 & 8.3 \\

ASTER~\cite{tpami19/aster/BGShi} 
& 74.2 & 78.8 & 61.3 & 65.9 & 75.9 & 70.6 & 76.9
& 96.2 & 94.6 & 97.6 & 87.4 & 88.7 & 93.4
& 81.66 & 19.1 \\

NRTR~\cite{icdar19/nrtr/FFSheng} 
& 67.9 & 42.4 & 66.5 & 73.6 & 66.4 & 77.2 & 78.3
& 97.8 & 96.8 & 98.1 & 88.9 & 93.3 & 94.4
& 80.12 & 44.3 \\

SAR~\cite{aaai19/sar/HLi} 
& 73.2 & 63.5 & 60.1 & 68.8 & 64.2 & 75.4 & 74.7
& 96.7 & 94.0 & 97.7 & 84.8 & 84.5 & 94.4
& 82.67 & 57.5 \\

RoScanner~\cite{eccv20/robustscanner/XYYue} 
& 79.4 & 68.1 & 70.5 & 79.6 & 71.6 & 82.5 & 80.8
& 97.7 & 95.8 & 98.5 & 88.2 & 90.1 & 97.6
& 84.65 & 48.0 \\

SRN~\cite{cvpr20/srn/DLYu} 
& 78.1 & 63.2 & 66.3 & 65.3 & 71.4 & 58.3 & 76.5
& 97.5 & 96.3 & 97.2 & 87.9 & 90.9 & 96.9
& 80.44 & 51.7 \\

SEED~\cite{cvpr20/seed/ZQiao} 
& 69.1 & 80.9 & 56.9 & 63.9 & 73.4 & 61.3 & 76.5
& 94.2 & 93.2 & 96.5 & 87.5 & 88.7 & 93.4
& 79.66 & 24.0 \\

VisionLAN~\cite{iccv21/visionlan/YXWang} 
& 79.6 & 71.4 & 67.9 & 73.7 & 76.1 & 73.9 & 79.1
& 97.1 & 95.8 & 98.2 & 88.6 & 91.2 & 96.2
& 83.75 & 32.9 \\

PIMNet~\cite{mm21/pimnet/ZQiao} 
& 80.3 & 79.8 & 68.4 & 75.9 & 77.8 & 68.3 & 80.8
& 98.4 & 97.8 & 98.9 & 89.6 & 94.3 & 97.9
& 85.24 & 24.2 \\

ABINet~\cite{cvpr21/abinet/SCFang} 
& 82.7 & 77.4 & 68.8 & 69.2 & 77.9 & 72.0 & 77.9
& 97.8 & 97.7 & 98.1 & 89.4 & 93.2 & 97.6
& 84.59 & 36.9 \\

SVTR~\cite{ijcai22/svtr/YKDu} 
& 79.3 & 69.4 & 67.1 & 68.8 & 75.8 & 76.8 & 77.0
& 96.7 & 96.7 & 98.2 & 88.6 & 91.3 & 96.5
& 83.27 & 18.1 \\

PARSeq~\cite{eccv22/parseq/Bautista} 
& 87.6 & 88.3 & 72.3 & 77.4 & 84.0 & 80.8 & 82.6
& 97.8 & 97.2 & 98.7 & 90.6 & 94.6 & 96.8
& 88.40 & 23.8 \\

MATRN~\cite{eccv22/matrn/Na} 
& 82.2 & 73.0 & 73.4 & 76.9 & 79.4 & 77.4 & 81.0
& 97.9 & \textbf{98.3} & 98.8 & 90.3 & 95.2 & 97.2
& 86.24 & 44.3 \\

MGP-STR~\cite{eccv22/mgpstr/PWang} 
& 85.2 & 83.7 & 72.6 & 75.1 & 79.8 & 71.1 & 83.1
& 97.1 & 97.8 & 97.9 & 89.6 & 95.2 & 96.9
& 86.54 & 148.0 \\

LPV~\cite{ijcai23/lpv/BQZhang} 
& 86.2 & 78.7 & 75.8 & 80.2 & 82.9 & 81.6 & 82.9
& 98.1 & 97.8 & 98.6 & 89.8 & 93.6 & 97.6
& 88.01 & 30.5 \\

MAERec~\cite{iccv23/maerec/QJiang} 
& 81.4 & 71.4 & 72.0 & 80.0 & 78.5 & 82.4 & 82.5
& 97.6 & 96.8 & 98.0 & 87.1 & 93.2 & 97.9
& 86.22 & 35.7 \\

LISTER~\cite{iccv23/lister/CXCheng} 
& 87.1 & 88.2 & 72.5 & 78.3 & 79.7 & 81.3 & 80.3
& 97.3 & 96.6 & 98.5 & 88.2 & 90.7 & 96.5
& 86.32 & 20.5 \\

CAM~\cite{pr24/cam/MKYang} 
& 85.4 & 89.0 & 72.0 & 75.4 & 84.0 & 74.8 & \underline{83.1}
& 96.6 & 96.1 & 98.2 & 89.0 & 93.5 & 96.2
& 87.18 & 58.7 \\

CDistNet~\cite{ijcv24/cdistnet/TLZheng} 
& 81.7 & 77.1 & 72.6 & 78.2 & 79.9 & 79.7 & 81.1
& 97.8 & 98.1 & 98.7 & 89.6 & 93.5 & 96.9
& 86.45 & 43.3 \\

BUSNet~\cite{aaai24/busnet/JJWei} 
& 83.0 & 82.3 & 70.8 & 77.9 & 78.8 & 71.2 & 82.6
& 97.8 & 98.1 & 98.3 & 90.2 & 95.3 & 96.5
& 86.39 & 32.1 \\

OTE~\cite{cvpr24/ote/JJXu} 
& 87.9 & 82.8 & 75.3 & 73.4 & 81.8 & 68.9 & 80.9
& 97.7 & 97.8 & 98.2 & 89.6 & 94.4 & 97.9
& 86.66 & 20.0 \\

IGTR~\cite{tpami25/igtr/YKDu} 
& 90.3 & 92.6 & 77.5 & 78.8 & 85.6 & 81.8 & 82.9
& 98.6 & \textbf{98.3} & 98.9 & 90.4 & 94.7 & 97.6
& 89.85 & 24.1 \\

CPPD~\cite{tpami25/cppd/YKDu} 
& 87.9 & 81.7 & 75.5 & 76.2 & 83.5 & 81.5 & 81.9
& 98.2 & 98.1 & 98.8 & \textbf{91.1} & 93.8 & 97.6
& 88.14 & 26.9 \\

IPAD~\cite{ijcv25/ipad/XMYang} 
& 85.2 & 83.3 & 72.1 & 78.4 & 81.4 & 73.7 & 82.2
& 98.1 & 97.7 & 99.0 & 90.8 & \underline{95.5} & 97.9
& 87.33 & 24.7 \\

SVTRv2~\cite{arxiv/svtrv2/YKDu} 
& 91.0 & 89.5 & \underline{78.7} & \textbf{81.3} & 85.9 & 85.1 & 82.3
& \underline{98.7} & 98.0 & \textbf{99.2} & 91.0 & 93.6 & \textbf{99.0}
& 90.30 & 21.0 \\

SMTR~\cite{aaai25/smtr/YKDu} 
& 90.5 & 92.6 & 75.5 & 80.0 & 84.9 & 85.2 & 82.6
& 98.4 & 98.2 & 98.8 & 90.0 & 93.3 & 97.6
& 88.93 & 15.8 \\

\hline
RISTER-T 
& 89.4 & 94.5 & 72.0 & 74.0 & 85.2 & 79.7 & 80.0
& 98.3 & 96.6 & 98.0 & 89.2 & 93.8 & 95.1
& 88.13 & 15.9 \\

RISTER-S 
& 92.5 & 96.0 & 76.7 & 77.8 & 87.8 & 84.6 & 81.9
& 98.4 & 97.8 & 98.6 & 90.0 & \underline{95.5} & 97.2
& 90.37 & 21.8 \\

RISTER-B 
& \underline{93.0} & \textbf{96.6} & 78.2 & 78.6 & \underline{89.1} & \underline{86.5} & 83.0
& 98.6 & \textbf{98.3} & 98.9 & \textbf{91.1} & 95.2 & \underline{98.3}
& \underline{91.18} & 32.8 \\

RISTER-L 
& \textbf{94.1} & \textbf{96.6} & \textbf{79.4} & \underline{80.6} & \textbf{90.3} & \textbf{87.4} & \textbf{83.7}
& \textbf{98.9} & 98.1 & \underline{99.0} & 90.6 & \textbf{95.7} & 97.6
& \textbf{92.46} & 38.6 \\

\hline
\end{tabular}
\label{tab:merged_avg}
\end{table*}

\subsection{Comparison with State-of-the-arts}
To evaluate the performance of our RISTER on general benchmarks, we compare it with existing and widely used STR methods. The results are shown in Table~\ref{tab:merged_avg}. As shown by the experimental results, our RISTER outperforms existing STR methods even on the standard 0° benchmarks. Moreover, it achieves superior performance on more challenging datasets, such as U14M-B, reaching state-of-the-art results. Specifically, RISTER-T achieves an average accuracy of 88.13\% while maintaining a relatively small number of parameters. Meanwhile, RISTER-S and RISTER-B, as medium-sized models in terms of parameter count, outperform existing STR methods on both CoB and U14M-B benchmarks. When the parameter size is further increased, the resulting RISTER-L becomes a powerful recognition model, achieving the highest accuracy on 8 out of 13 test benchmarks, thereby setting new state-of-the-art performance. Meanwhile, it still maintains a parameter count below 40M. The superiority of RISTER on these 0° test benchmarks can be attributed to two main factors:
(1) the well-designed four-stage RELG architecture, which enables efficient feature extraction while maintaining a relatively small number of parameters; and
(2) the integration of rotation invariance into the model, which allows RISTER to learn effectively without relying on rotated training samples. This not only improves the efficiency of training data utilization but also enhances the aggregation of key features, thereby strengthening the model’s robustness.


\section{Conclusion}
In this work, we propose RISTER, a system-level, rotation-invariant framework for STR. RISTER provides theoretical guarantees for multi-oriented text recognition and addresses common limitations of existing methods, such as increased computational cost and heavy reliance on data augmentation. Experimental results demonstrate that RISTER consistently achieves superior performance across both multi-oriented benchmarks and horizontally oriented samples. In future work, we plan to explore extending RISTER-like rotation-invariant frameworks to non-square input settings to support applications such as mathematical expression recognition and long-text recognition.


\section*{Acknowledgements}
This work was supported by the National Natural Science Foundation of China (62302532), Guangdong Basic and Applied Basic Research Foundation (2025A15-15011224), Shenzhen Science and Technology Program (KQTD202211010935590-18, SYSRD20250529113401002), Open Research Fund of The State Key Laboratory of Multimodal Artificial Intelligence Systems (MAIS2025052),  Basic Research Program of Jiangsu (BK20251441), and sponsored by CCF-Baidu Open Fund (OF202519). The model is implemented based on PaddlePaddle.

%
%
\clearpage
\bibliographystyle{splncs04}
\bibliography{main}
\end{document}